\PassOptionsToPackage{unicode}{hyperref}
\PassOptionsToPackage{hyphens}{url}
\PassOptionsToPackage{numbers,compress}{natbib}
\documentclass[
]{article}
\usepackage[preprint]{neurips_2025}
\usepackage{xcolor}
\usepackage{amsmath,amssymb}
\usepackage{iftex}
\ifPDFTeX
  \usepackage[T1]{fontenc}
  \usepackage[utf8]{inputenc}
  \usepackage{textcomp} 
\else 
  \usepackage{unicode-math} 
  \defaultfontfeatures{Scale=MatchLowercase}
  \defaultfontfeatures[\rmfamily]{Ligatures=TeX,Scale=1}
\fi
\usepackage{lmodern}
\ifPDFTeX\else
\fi
\IfFileExists{upquote.sty}{\usepackage{upquote}}{}
\IfFileExists{microtype.sty}{
  \usepackage[]{microtype}
  \UseMicrotypeSet[protrusion]{basicmath} 
}{}
\makeatletter
\@ifundefined{KOMAClassName}{
  \IfFileExists{parskip.sty}{%
    \usepackage{parskip}
  }{
    \setlength{\parindent}{0pt}
    \setlength{\parskip}{6pt plus 2pt minus 1pt}}
}{
  \KOMAoptions{parskip=half}}
\makeatother
\usepackage{longtable,booktabs,array}
\usepackage{calc} 
\usepackage{etoolbox}
\makeatletter
\patchcmd\longtable{\par}{\if@noskipsec\mbox{}\fi\par}{}{}
\makeatother
\IfFileExists{footnotehyper.sty}{\usepackage{footnotehyper}}{\usepackage{footnote}}
\makesavenoteenv{longtable}
\usepackage{graphicx}
\usepackage{placeins}
\makeatletter
\newsavebox\pandoc@box
\newcommand*\pandocbounded[1]{
  \sbox\pandoc@box{#1}%
  \Gscale@div\@tempa{\textheight}{\dimexpr\ht\pandoc@box+\dp\pandoc@box\relax}%
  \Gscale@div\@tempb{\linewidth}{\wd\pandoc@box}%
  \ifdim\@tempb\p@<\@tempa\p@\let\@tempa\@tempb\fi
  \ifdim\@tempa\p@<\p@\scalebox{\@tempa}{\usebox\pandoc@box}%
  \else\usebox{\pandoc@box}%
  \fi%
}
\def\fps@figure{htbp}
\makeatother
\usepackage{bookmark}
\IfFileExists{xurl.sty}{\usepackage{xurl}}{} 
\hypersetup{
  pdftitle={Brain-Grounded Axes for Reading and Steering LLM States},
  pdfauthor={Sandro Andric},
  hidelinks,
  pdfcreator={LaTeX via pandoc}}

\title{Brain-Grounded Axes for Reading and Steering LLM States}
\author{Sandro Andric\\\texttt{sandro.andric@nyu.edu}}
\date{}

\begin{document}
\maketitle
\begin{abstract}
Interpretability methods for large language models (LLMs) typically
derive directions from textual supervision, which can lack external
grounding. We propose using human brain activity not as a training
signal but as a coordinate system for reading and steering LLM states.
Using the SMN4Lang MEG dataset, we construct a word-level brain atlas of
phase-locking value (PLV) patterns and extract latent axes via ICA. We
validate axes with independent lexica and NER-based labels
(POS/log-frequency used as sanity checks), then train lightweight
adapters that map LLM hidden states to these brain axes without
fine-tuning the LLM. Steering along the resulting brain-derived
directions yields a robust lexical (frequency-linked) axis in a mid
TinyLlama layer, surviving perplexity-matched controls, and a
brain-vs-text probe comparison shows larger log-frequency shifts
(relative to the text probe) with lower perplexity for the brain axis. A
function/content axis (axis 13) shows consistent steering in TinyLlama,
Qwen2-0.5B, and GPT-2, with PPL-matched text-level corroboration.
Layer-4 effects in TinyLlama are large but inconsistent, so we treat
them as secondary (Appendix). Axis structure is stable when the atlas is
rebuilt without GPT embedding-change features or with word2vec
embeddings (\(|r|=0.64\text{--}0.95\) across matched axes), reducing circularity
concerns. Exploratory fMRI anchoring suggests potential alignment for
embedding change and log frequency, but effects are sensitive to
hemodynamic modeling assumptions and are treated as population-level
evidence only. These results support a new interface:
neurophysiology-grounded axes provide interpretable and controllable
handles for LLM behavior.
\end{abstract}

\section{Introduction}\label{introduction}

Most LLM interpretability discovers directions using text-derived labels
or optimization objectives. Such directions can be effective yet are
rarely grounded in external measurements of human cognition. We ask
whether brain activity can define a stable, externally grounded
coordinate system that enables both reading and steering of LLM
representations. Our key idea is to compute a brain-derived atlas of
semantic activity and use its axes as the target space for LLM
projections. We do not optimize the LLM; instead we train a lightweight
adapter and test whether steering along those axes produces reliable,
interpretable shifts in generated text.

Contributions: - We build a MEG-derived word-level brain atlas of PLV
connectivity and extract latent axes with stability/robustness checks. -
We show that a lightweight adapter can map LLM hidden states into this
brain axis space without fine-tuning. - We demonstrate robust steering
for a frequency-linked axis (TinyLlama L11) and cross-model steering for
a function/content axis (GPT‑2, Qwen‑0.5B, TinyLlama), including
PPL-matched and brain‑vs‑text probe controls.

\section{Related Work}\label{related-work}

Prior work aligns neural data with language models for encoding/decoding
and representation comparisons
\citep{schrimpf2021neural,jain2018context,gauthier2019linking,caucheteux2022converge},
uses brain data to interpret NLP models \citep{toneva2019interpreting},
and demonstrates semantic reconstruction from brain recordings
\citep{tang2023semantic}. Recent steering methods use activation-addition
and contrastive vectors \citep{turner2023actadd,rimsky2023steering}. Our
contribution differs in two ways:
(1) we derive axes from MEG connectivity geometry rather than textual
supervision, and (2) we treat brain axes as a fixed interface for
reading and steering, not as a reward for optimization.

\section{Dataset}\label{dataset}

We use SMN4Lang (OpenNeuro ds004078 v1.2.1)
\citep{wang2022smn4lang,openneuro_ds004078}, a synchronized MEG/fMRI
dataset of naturalistic story listening. We use
preprocessed MEG sensor-level data and word-level timing annotations.
The dataset includes 12 subjects and 60 stories.

\section{Methods}\label{methods}

\subsection{MEG PLV Atlas}\label{meg-plv-atlas}

We compute phase-locking value (PLV) in theta band (4--8 Hz) over
sliding windows (2.0 s length, 0.5 s step) on gradiometer channels
\citep{lachaux1999measuring}. For each window, we compute PLV
connectivity and
compress the edge vector using PCA to 128 dimensions (edge-PCA).
Filtering and analytic phase extraction are implemented with MNE-Python
\citep{gramfort2013mne}. This yields a time-indexed PLV state vector per
run.

\subsection{Word-Level Brain Atlas}\label{word-level-brain-atlas}

We align PLV windows to word-level timing annotations and build a
word-level atlas using ridge regression \citep{hoerl1970ridge} with
features: embedding change (GPT), word log-frequency, and POS id. Lags
of 0.0, 0.5, and 1.0 s are included. The model predicts PLV-PCA states
per window; each window is assigned to the most recent word onset, and
we average predicted PLV vectors by word type to obtain a word-level
atlas per subject. We use predicted PLV states (rather than single-trial
PLV averages) to improve signal-to-noise and stabilize word-level
estimates. Subject-level atlases are then averaged across subjects to
form a cross-subject atlas. To test circularity, we rebuild the atlas
with embedding change removed (logfreq+POS only) and with word2vec
embeddings \citep{mikolov2013word2vec}, then align axes by correlation.
Atlas predictions are generated out-of-fold by run (5-fold): for each
fold, models are trained on a subset of runs and used to predict held-out
runs, and the word atlas is built from these OOF predictions (full-data
weights are saved for reference).

\subsection{Axis Discovery}\label{axis-discovery}

We apply ICA to the averaged word atlas and obtain latent axes
\citep{hyvarinen2000independent}. We fix the number of components to 20
a priori
for stability across runs. We validate axes using independent label
sources:

\begin{itemize}
\item
  CKIP POS/NER for function/content, noun/verb, and animate labels
  \citep{ckiptagger} (temporal heuristics used only for exploratory
  checks)
\item
  MELD-SCH concreteness \citep{tsang2018meld,xu2020concreteness} and
  Chinese valence/arousal lexica \citep{yu2016cvaw}
\end{itemize}

Permutation tests confirm axis-label alignment. Axis labels in the paper
(e.g., function/content, animacy) are assigned by the strongest
atlas‑label association and are not selected using steering outcomes.
Axis IDs refer to fixed ICA component indices; when comparing across
layers or models, we align axis direction by correlation and report
signed effects after alignment (orientation is arbitrary).

\subsection{LLM Adapter}\label{llm-adapter}

We extract TinyLlama word-level hidden states for all stories
\citep{tinyllama2024} and map them to brain axes using ridge regression
(standardized, alpha chosen by CV). This yields an adapter \(f(h)=Wh+b\)
that predicts axis scores per word.

\subsection{Steering}\label{steering}

We steer TinyLlama by adding a normalized axis vector to hidden states
at selected layers during generation. Concretely, for axis \(k\) we use
the corresponding adapter weight vector \(W_k\) (scaled back if
standardization is used), L2-normalize it, and add
\(\alpha W_k/\|W_k\|\) to the hidden state at the chosen layer for all
positions in each forward pass (prompt and generated tokens). We test
strengths \(\{-5,-2,-1,0,1,2,5\}\) over 50 prompts with 4 samples per
strength (256 tokens each) using batched generation, and evaluate:

\begin{itemize}
\item
  Adapter-score shift (pos vs neg, t-test + permutation)
\item
  Strength correlation
\item
  Fluency (perplexity) shift
\item
  Text-level POS/NER metrics
\end{itemize}

We correct for multiple comparisons across layers and axes with FDR
\citep{benjamini1995controlling}.

\section{Results}\label{results}

\subsection{Axis Validation}\label{axis-validation}

Word-level validation confirms strong axis-label alignment on non-POS
lexica:

\begin{itemize}
\item
  Animate vs inanimate (confound-controlled): axis 2, residualized
  \(d=0.53\) (logfreq+surprisal+length), matched \(d=0.70\) with CI
  {[}0.53, 0.86{]} (n=330)
\item
  Lexical frequency: axis 15, \(r=0.51\) with logfreq (n=13,632)
\end{itemize}

Lexica validation: concreteness (axis 2, \(r=-0.128\), \(p=0.001\)),
valence (axis 15, \(r=0.114\), \(p=0.001\)), arousal (axis 15,
\(r=0.084\), \(p=0.001\)). Axis 15 is strongly frequency-linked, so
valence/arousal effects are treated as secondary. POS-derived
validations (function/content, noun/verb) are reported as sanity checks
only (Appendix).

Confound control shows that animacy is not driven by
logfreq/surprisal/length: axis 2 remains significant after
residualization (\(d=0.53\), \(p=7.1\times10^{-5}\)), while axis 15
attenuates (residualized \(d=0.13\), \(p=0.25\)).

Out-of-fold (OOF) atlas predictions preserve the axis geometry: the
best-match correlations for axes 13/19/15/2 are \(|r|=0.82\text{--}0.97\)
relative to the base atlas. For reproducibility we match OOF axes to the
original indices by correlation (2\(\rightarrow\)3, 13\(\rightarrow\)11,
15\(\rightarrow\)5, 19\(\rightarrow\)12) and keep the base labels
throughout the paper.

Leakage-robust control: in a wPLI run (4 subjects, runs 10--20, decim=5),
the lexical-frequency axis remains robust (\(|r|=0.744\)), while other
axes show weaker correspondence (\(|r|\approx0.35\text{--}0.48\)),
suggesting non-lexical axes are less stable under leakage-robust
connectivity at this reduced-data setting.

\subsection{Embedding-Change
Ablations}\label{embedding-change-ablations}

Rebuilding the atlas without GPT embedding-change features (logfreq+POS
only) or with word2vec embeddings yields axes highly correlated with the
base atlas (\(|r|=0.82\text{--}0.95\) for axes 13/19/15/2 under no-embedding,
\(|r|=0.64\text{--}0.93\) under word2vec), indicating the axis geometry is not
driven by GPT features. Dropping POS features substantially alters the
POS-linked axes (\(|r|<0.3\)), confirming these labels are not
independent; we therefore do not treat POS-derived axis validation as
primary evidence. Dropping log-frequency features markedly reduces
lexical-axis alignment (axis 15 correlation drops to \(|r|=0.19\)), so
we treat this axis as supervised rather than emergent.

\subsection{Axis Reliability and Cross-Subject
Generalization}\label{axis-reliability-and-cross-subject-generalization}

We ran a rigorous reliability analysis with bootstrap confidence
intervals and permutation tests across 60 axis-lexicon pairs. After FDR
correction, 25/60 tests remain significant (expected 3 by chance). The
strongest effects include concreteness (axis 2, \(r=-0.128\) with CI
{[}-0.162, -0.092{]}), valence (axis 15, \(r=0.114\) with CI {[}0.082,
0.141{]}), and arousal (axis 15, \(r=0.084\) with CI {[}0.056,
0.114{]}). Partial correlations controlling for word length show
negligible change for concreteness; axis 15 remains strongly tied to log
frequency (\(r=0.51\)).

Cross-subject validation (odd vs even splits) shows 12 axis-dimension
pairs replicating with \(|r|>0.05\). We align axes by correlation and
report absolute values to avoid sign ambiguity (ICA orientation is
arbitrary). Examples include the matched concreteness axis (test
\(|r|=0.079 \pm 0.014\)), valence axis (test \(|r|=0.093 \pm 0.023\)),
and arousal axis (test \(|r|=0.083 \pm 0.005\)). This supports stable
semantic structure across subjects.

\subsection{Adapter Transfer to LLM}\label{adapter-transfer-to-llm}

The TinyLlama adapter generalizes to held-out words (9,149 matched
words):

\begin{itemize}
\item
  Axis 13: \(r=0.238\) (\(p<10^{-24}\))
\item
  Axis 19: \(r=0.499\) (\(p<10^{-115}\))
\item
  Axis 15: \(r=0.461\) (\(p<10^{-96}\))
\item
  Axis 2: \(r=0.624\) (\(p<10^{-197}\))
\end{itemize}

\textbf{Cross-model transfer (Qwen2-0.5B).} Using the same atlas, a
lightweight adapter trained on Qwen2-0.5B hidden states \citep{qwen22024}
also predicts brain axes on held-out words (9,149 matched): axis
13 \(r=0.264\), axis 19 \(r=0.528\), axis 15 \(r=0.435\), axis 2
\(r=0.637\) (all \(p\ll 10^{-30}\)). This supports the ``interface''
framing beyond a single LLM. Steering in Qwen2-0.5B shows significant
adapter-score shifts at layer 12 for axes 13/15/2 (perm \(p\leq0.002\)),
with axis 19 weaker (perm \(p=0.029\)), and a strong axis 13 effect at
layer 4. GPT-2 \citep{radford2019gpt2} shows a consistent axis 13 effect
across layers (best L6 \(d=0.30\), perm \(p=0.001\)), while axis 15 is
null in GPT-2. These results provide cross-model steering evidence for
axis 13; axis 15 appears model-dependent (present in TinyLlama and
Qwen‑0.5B, absent in GPT‑2).

\subsection{Steering (Primary
Evidence)}\label{steering-primary-evidence}

Primary result: a lexical frequency-linked (supervised) axis (axis 15)
at TinyLlama layer 11 shows robust steering with PPL-matched
confirmation and a brain‑vs‑text probe efficiency advantage. We treat
independent text-level metrics as primary evidence and use adapter-score
shifts only as a manipulation check. Layers tested: 4, 11, 20
(TinyLlama). In the batched 50-prompt run, axis 15 yields \(d=0.545\)
(perm \(p=0.001\)). Under PPL-matched controls with the same 50-prompt
setup, the effect remains (\(d=0.571\), perm \(p=0.001\)). A
brain-vs-text probe comparison at the same layer shows that the brain
axis produces large log-frequency shifts with lower perplexity
(\(d=0.9246\); PPL \(d=-0.2183\)), while a text-only logfreq probe
shifts in the opposite direction and increases PPL (\(d=-0.3021\); PPL
\(d=0.4549\)). We therefore treat log-frequency as the primary
independent outcome for axis 15 and use adapter-score shifts as a
manipulation check.

To benchmark against standard representation engineering, we computed an
Activation Addition (ActAdd) baseline using the mean difference between
the top/bottom 50 log-frequency words at TinyLlama layer 11. As
expected, this supervised ActAdd vector produced a larger log-frequency
shift (\(d=1.6666\), perm \(p=0.001\)) \citep{turner2023actadd}. The
brain
axis remained competitive (\(d=0.9246\)) and differed in two key ways:
it significantly improved fluency (PPL \(d=-0.2183\), perm \(p=0.001\)),
while ActAdd had no significant PPL effect (PPL \(d=-0.0740\), perm
\(p=0.237\)), and the directions are nearly orthogonal (cosine
similarity \(0.0104\)).

Across models, axis 13 (function/content) shows consistent steering:
TinyLlama L11 (\(|d|=0.208\), perm \(p=0.001\)), Qwen-0.5B L4
(\(d=0.436\), perm \(p=0.001\)), and GPT-2 L6 (\(d=0.300\), perm
\(p=0.001\)). Perplexity-matched controls for axis 13 in TinyLlama L11
and Qwen L4 preserve the effect and yield strong text-level
function/content shifts. Other layers and axes show weaker or
inconsistent effects; full summaries appear in Appendix (Table 1, Fig.
1).

TinyLlama layer 4 shows large raw shifts but sign flips and lower
prompt-level consistency across axes; we therefore keep layer-4 effects
as secondary evidence.

\protect\phantomsection\label{tab:best_layers}
\begin{longtable}[]{@{}llllll@{}}
\toprule\noalign{}
model & axis & label & selected layer & d & perm\_p \\
\midrule\noalign{}
\endhead
\bottomrule\noalign{}
\endlastfoot
Qwen2-0.5B & 13 & function\_content & 4 & 0.436 & 0.000999 \\
Qwen2-0.5B & 15 & lexical\_freq & 12 & 0.380 & 0.000999 \\
Qwen2-0.5B & 19 & noun\_verb & 16 & 0.212 & 0.000999 \\
Qwen2-0.5B & 2 & animacy & 12 & 0.181 & 0.002 \\
GPT-2 & 13 & function\_content & 6 & 0.300 & 0.000999 \\
GPT-2 & 15 & lexical\_freq & 6 & -0.032 & 0.57 \\
GPT-2 & 19 & noun\_verb & 3 & 0.140 & 0.018 \\
GPT-2 & 2 & animacy & 12 & 0.109 & 0.0509 \\
TinyLlama & 13 & function\_content & 11 & -0.208 & 0.000999 \\
TinyLlama & 15 & lexical\_freq & 11 & 0.545 & 0.000999 \\
TinyLlama & 19 & noun\_verb & 11 & 0.316 & 0.000999 \\
TinyLlama & 2 & animacy & 11 & 0.194 & 0.002 \\
\end{longtable}

Selected layer per model and axis. For TinyLlama we report the primary
layer (11) to avoid unstable layer‑4 effects.

As a baseline, a matched random-direction steering run at layer 11 (same
prompts/strengths/samples/tokens) yields no effect (axis 15 random:
\(p=0.88\), \(d=-0.02\)), indicating that arbitrary directions do not
produce the observed shifts.

\begin{figure}[!ht]
\centering
\pandocbounded{\includegraphics[keepaspectratio,alt={Layer 11 axis 15 steering curve (TinyLlama; batched 50 prompts). Adapter-score means are plotted per strength.}]{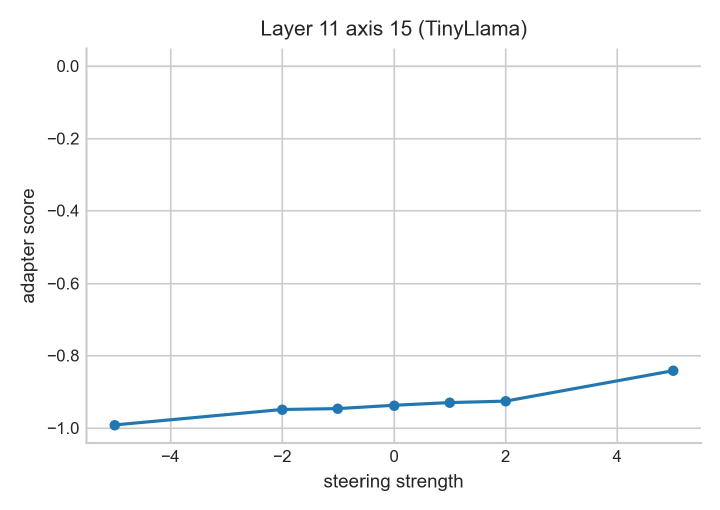}}
\caption{Layer 11 axis 15 steering curve (TinyLlama; batched 50
prompts). Adapter-score means are plotted per
strength.}\label{fig:layer11_axis15}
\end{figure}

\begin{figure}[!ht]
\centering
\pandocbounded{\includegraphics[keepaspectratio,alt={Efficiency comparison for brain axis, text probe, and ActAdd steering (TinyLlama L11). Brain-axis steering yields a large log-frequency shift with improved perplexity; ActAdd yields a larger shift with no significant PPL change.}]{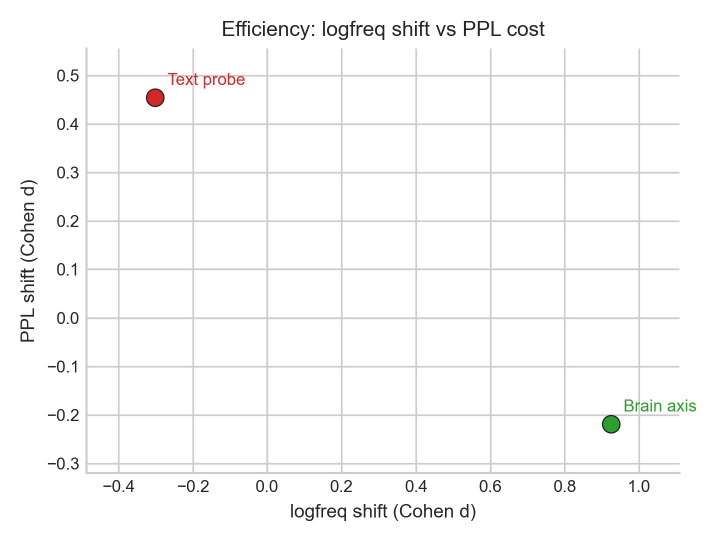}}
\caption{Efficiency comparison for brain axis, text probe, and ActAdd
steering (TinyLlama L11). Brain-axis steering yields a large
log-frequency shift with improved perplexity; ActAdd yields a larger
shift with no significant PPL change.}\label{fig:brain_vs_text_probe}
\end{figure}

\begin{figure}[!ht]
\centering
\pandocbounded{\includegraphics[keepaspectratio,alt={Selected-layer steering effects by model and axis (cell color encodes Cohen d; numbers show d values).}]{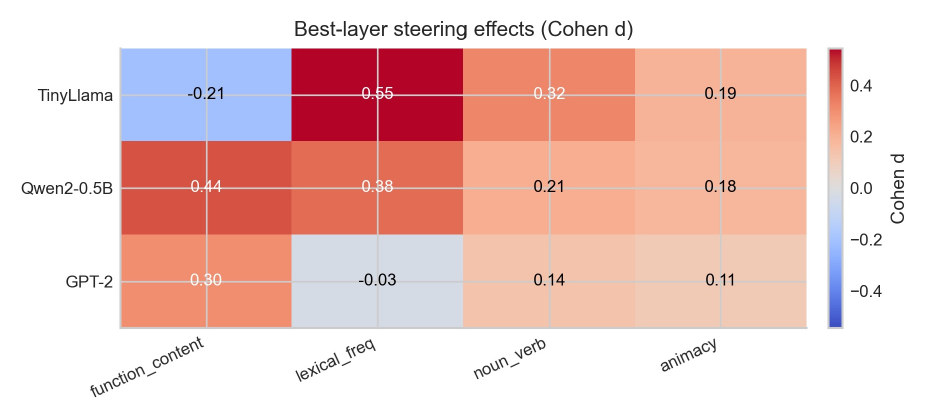}}
\caption{Selected-layer steering effects by model and axis (cell color
encodes Cohen d; numbers show d values).}\label{fig:best_layers_dotplot}
\end{figure}

\FloatBarrier

\subsection{Perplexity-Matched
Controls}\label{perplexity-matched-controls}

The primary effect (layer 11, axis 15) remains significant under
perplexity matching (see above). We also ran PPL-matched controls for
axis 13 in TinyLlama L11 and Qwen L4; both preserve adapter-score
effects and show strong text-level function/content shifts (Appendix).

\subsection{External Text-Level
Validation}\label{external-text-level-validation}

POS/NER metrics show partial external corroboration:

\begin{itemize}
\item
  Log-frequency mean: significant at layer 11 axis 15 (perm \(p=0.001\);
  \(d=0.765\))
\item
  Function/content ratio: significant at layer 11 axis 13 (perm
  \(p=0.001\); \(d=0.739\)) and under PPL-matched controls in TinyLlama
  L11 and Qwen L4 (perm \(p=0.001\))
\item
  Animate rate: significant at layer 11 axis 2 (perm \(p=0.001\);
  \(d=0.169\))
\item
  Noun ratio: significant at layer 11 axis 19 (perm \(p=0.001\);
  \(d=0.817\)) and weaker at layer 4 axis 19 (perm \(p=0.015\);
  \(d=-0.156\))
\end{itemize}

These text-level shifts are treated as secondary evidence because
adapter-score effects are the primary criterion; we therefore report
both to avoid measurement-within-the-loop concerns. POS-based text
metrics (function/content, noun ratio) are not independent because POS
tags are used in atlas construction and are treated as sanity checks
only.

\subsection{fMRI Anchoring
(Exploratory)}\label{fmri-anchoring-exploratory}

We ran an encoding-based fMRI anchoring test (ridge) and varied the HRF.
Using a 4~s peak HRF, pooled permutation tests across subjects and runs
(n=240) show weak but significant effects for embedding change (observed
mean abs corr 0.0888 vs null 0.0718, perm \(p=0.001\)) and log frequency
(0.0815 vs 0.0725, perm \(p=0.009\)). POS id is not significant (perm
\(p=0.152\)). Per-subject effects are inconsistent, so we treat fMRI
anchoring as population-level evidence only; sensitivity analyses are in
Appendix.

\section{Discussion}\label{discussion}

Brain-derived axes can steer LLM behavior without LLM fine-tuning.
Strongest evidence appears for a lexical (frequency-linked) axis in a
mid layer; the animacy axis (axis 2) is robust in the atlas but is not
the strongest steering effect. Fluency-controlled analyses confirm the
primary effect is not driven by perplexity shifts. External text metrics
provide the primary independent validation, while adapter-score shifts
are treated as manipulation checks.

A key concern is circularity: do brain-derived axes simply recover text
statistics? The brain-vs-text probe comparison argues against this. The
brain axis shifts log-frequency in the intended direction with lower
perplexity cost (\(d=0.9246\); PPL \(d=-0.2183\)), while a text-only
logfreq probe shifts in the opposite direction and increases perplexity
(\(d=-0.3021\); PPL \(d=0.4549\)). This suggests that mapping through
neurophysiology can act as a functional filter, isolating causal
structure that a direct text probe does not capture.

Baseline comparisons also reveal distinct geometric mechanisms. The
frequency axis (axis 15) is nearly orthogonal to the supervised ActAdd
vector (cosine similarity \(0.0104\)), yet both steer log-frequency.
This suggests multiple frequency-related subspaces: a direct statistical
direction (ActAdd) and a more naturalistic direction derived from the
brain atlas. Steering along the brain-derived path significantly
improves perplexity, while ActAdd yields no significant PPL change.

\section{Limitations}\label{limitations}

The atlas is sensor-space and may include field spread
\citep{nolte2004identifying,colclough2015symmetric}. Text-level metrics
are limited by
automatic tagging accuracy; we treat them as primary but imperfect
evidence and use adapter-score shifts as manipulation checks alongside
PPL controls. POS-derived validations are not independent because POS
features are used in atlas construction; we report them as sanity checks
only. Some axes show sign reversals across layers, reflecting
orientation ambiguity or layer-specific encoding. The strongest steering
axis is frequency-linked (supervised), so semantic purity is limited;
future work should disentangle lexical confounds. Cross-model transfer
is shown for adapter readout (Qwen2-0.5B), and cross-model steering is
supported for axis 13 in Qwen2-0.5B and GPT-2, but axis 15 is
model-dependent (present in TinyLlama/Qwen-0.5B, absent in GPT-2). fMRI
anchoring yields weak global with inconsistent per-subject reliability.
We do not claim exclusivity; other axes and layers may be steerable but
did not meet our strict robustness criteria here.

\section{Ethics}\label{ethics}

The work uses public, de-identified neuroimaging data. Steering results
are modest and should not be interpreted as mind-reading or behavioral
prediction of individuals.

\section{Reproducibility}\label{reproducibility}

Code, configs, and minimal reproduction instructions are available at:
\url{https://github.com/sandroandric/Brain-Grounded-Axes-for-Reading-and-Steering-LLM-States.git}.
The repository documents required inputs and the command-line steps to
reproduce all reported results.

\section{Supplementary Steering
Analyses}\label{supplementary-steering-analyses}

\protect\phantomsection\label{tab:steering_summary}
\begin{longtable}[]{@{}llrr@{}}
\toprule\noalign{}
Layer & Axis & \(d\) & perm \(p\) \\
\midrule\noalign{}
\endhead
\bottomrule\noalign{}
\endlastfoot
4 & function\_content & 0.1922 & 0.001998 \\
4 & lexical\_freq & -0.8490 & 0.001 \\
4 & noun\_verb & -0.2726 & 0.001 \\
4 & animacy & -0.1227 & 0.03397 \\
11 & function\_content & -0.2079 & 0.001 \\
11 & lexical\_freq & 0.5450 & 0.001 \\
11 & noun\_verb & 0.3163 & 0.001 \\
11 & animacy & 0.1936 & 0.001998 \\
20 & function\_content & -0.1128 & 0.04296 \\
20 & lexical\_freq & 0.2580 & 0.001 \\
20 & noun\_verb & -0.0106 & 0.845 \\
20 & animacy & 0.1340 & 0.01898 \\
\end{longtable}

Full steering summary across layers and axes (TinyLlama, batched
50-prompt run). perm \(p\) is from permutation tests on adapter-score
shift; signs are reported after axis matching (orientation is
arbitrary).

\subsection{Appendix Baselines}\label{appendix-baselines}

\begin{longtable}{@{}p{0.52\linewidth}rrrr@{}}
\toprule
Baseline (TinyLlama L11, 50 prompts) & logfreq $d$ & logfreq $p$ & PPL $d$ & PPL $p$ \\
\midrule
\endhead
\bottomrule
\endlastfoot
Random dir. (axis 15) & -0.02 & 0.88 & n/a & n/a \\
ActAdd (top/bot 50 logfreq) & 1.6666 & 0.001 & -0.0740 & 0.237 \\
Brain axis (axis 15) & 0.9246 & 0.001 & -0.2183 & 0.001 \\
\end{longtable}

\begin{figure}[!ht]
\centering
\pandocbounded{\includegraphics[keepaspectratio,alt={Layer 4 steering effects for the function/content axis (left) and lexical frequency axis (right), aligned to the layer-11 axis orientation. Plots show adapter-score shift across strengths.}]{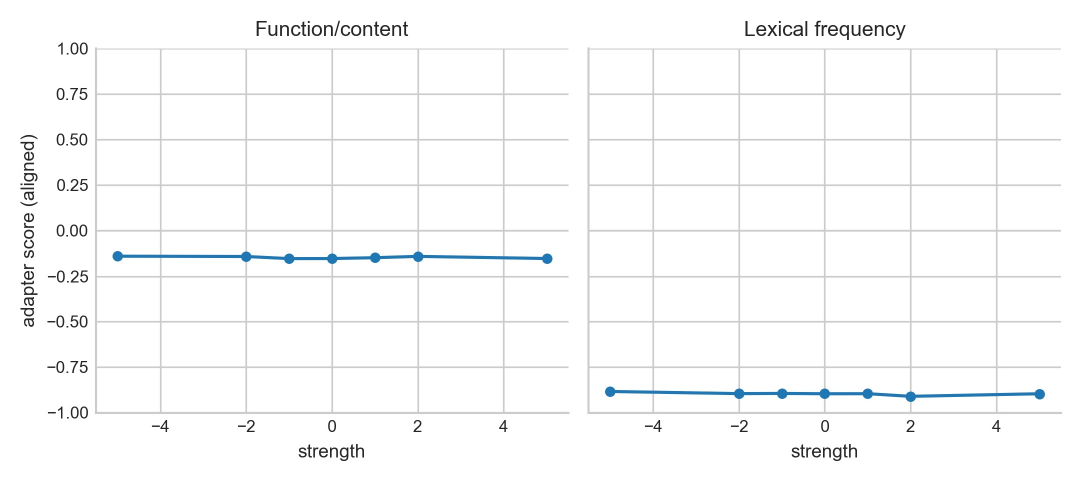}}
\caption{Layer 4 steering effects for the function/content axis (left)
and lexical frequency axis (right), aligned to the layer-11 axis
orientation. Plots show adapter-score shift across
strengths.}\label{fig:steering_layer4}
\end{figure}

Perplexity-matched checks for secondary effects:

\begin{itemize}
\item
  TinyLlama L11, axis 13: adapter shift \(d=-0.2355\) (perm
  \(p=0.001\)); function/content ratio \(d=0.6086\) (perm \(p=0.001\)).
\item
  Qwen L4, axis 13: adapter shift \(d=0.4364\) (perm \(p=0.001\));
  function/content ratio \(d=1.0620\) (perm \(p=0.001\)).
\end{itemize}

\section{Supplementary fMRI Anchoring
Sensitivity}\label{supplementary-fmri-anchoring-sensitivity}

HRF-off anchoring yields significant pooled effects (embedding change
\(p=0.001\), logfreq \(p=0.001\), pos\_id \(p=0.006\)) but with smaller
absolute correlations. The default 6~s peak HRF shows only marginal
pooled effects (embedding change \(p=0.071\), logfreq \(p=0.082\),
pos\_id \(p=0.039\)). These variants reinforce that fMRI anchoring is
weak and sensitive to HRF assumptions.

\protect\phantomsection\label{refs}
\bibliographystyle{plainnat}
\bibliography{references}

\end{document}